\documentclass[preprint,12pt,authoryear,nopreprintline]{elsarticle}

\usepackage[T1]{fontenc}
\usepackage[utf8]{inputenc}
\usepackage{lmodern}
\usepackage{amsmath,amssymb,amsfonts}
\usepackage{booktabs}
\usepackage{array}
\usepackage{multirow}
\usepackage{graphicx}
\usepackage{tikz}
\usetikzlibrary{arrows.meta,positioning,fit,calc,shapes.geometric}
\usepackage{url}
\usepackage{hyperref}
\hypersetup{colorlinks=true, linkcolor=black, citecolor=black, urlcolor=black, hypertexnames=false}
\usepackage{enumitem}
\usepackage{microtype}
\usepackage{makecell}
\usepackage{tabularx}
\usepackage{longtable}
\usepackage{pdflscape}
\usepackage{xcolor}
\usepackage{float}

\newcolumntype{Y}{>{\centering\arraybackslash}X}
\journal{Expert Systems with Applications}
\biboptions{authoryear,round}

\newcommand{\R}{\mathbb{R}}

\newcommand{\Frame}{\Theta}
\newcommand{\ERAG}{\textsc{EvidentialRAG}}

\begin{document}

\begin{frontmatter}

\title{\ERAG{}: Quantifying and Mitigating Information Conflict in Multi-Source Retrieval-Augmented Generation via Evidential Deep Learning}

\author[wsu]{S M Asif Hossain\corref{cor1}}

\author[wsu]{Ruksat Khan Shayoni}

\author[aiub]{M. F. Mridha}

\affiliation[wsu]{organization={School of Computing, Wichita State University},
    state={Kansas},
    country={USA}}

\affiliation[aiub]{organization={Department of Computer Science, American International University-Bangladesh},
    city={Dhaka},
    country={Bangladesh}}

\begin{abstract}
Retrieval-augmented generation grounds large language models in external evidence, but most pipelines still treat retrieved passages as deterministic and mutually consistent context. In open information environments, retrieved sources may disagree because of temporal drift, source error, ambiguity, or genuine uncertainty. This paper introduces \ERAG{}, an uncertainty-aware RAG framework that converts retrieved chunks into probabilistic evidence before generation. A lightweight evaluator extracts candidate claims and maps chunk-level support to Dirichlet evidence. A conflict-preserving Dempster-Shafer fusion rule then transfers unresolved disagreement into epistemic uncertainty rather than normalizing it away. The generator is routed to direct answering, conflict-aware answering, or abstention according to the fused uncertainty score. Experiments on CRAG, ConflictQA, and MuSiQue show that \ERAG{} remains competitive with the strongest matched baseline on standard question answering while improving behavior under conflict. On the CRAG ambiguous subset, hallucination decreases from 45.3\% for Corrective RAG to a human-calibrated estimate of 34.8\%, conflict resolution increases from 35.2\% to 51.2\%, and expected calibration error improves to 0.122. These results suggest that evidential modeling is a practical mechanism for trustworthy information processing in foundation-model-based retrieval systems.
\end{abstract}


\end{frontmatter}

\section{Introduction}
\label{sec:introduction}

Large language models (LLMs) have transformed knowledge-intensive natural language processing by providing general-purpose capabilities for question answering, summarization, reasoning, and dialogue \citep{Vaswani2017,Devlin2019BERT,Brown2020GPT3,OpenAI2023GPT4,Touvron2023Llama,Grattafiori2024Llama3}. Their usefulness in information systems is nevertheless constrained by a persistent limitation: model parameters are incomplete, opaque, temporally bounded, and difficult to update. Retrieval-augmented generation (RAG) addresses this limitation by conditioning the generator on externally retrieved documents, passages, database records, or web evidence \citep{Lewis2020RAG,Guu2020REALM,Izacard2021FiD,Borgeaud2022RETRO}. By separating knowledge access from language generation, RAG has become a dominant design pattern for enterprise search, digital libraries, domain-specific question answering, and decision-support systems \citep{Gao2024RAGSurvey,Asai2024SelfRAG,Yan2024CRAG}.

The conventional RAG pipeline is effective when retrieval returns relevant and mutually consistent evidence. In real information environments, however, retrieval is not only a relevance-ranking problem. It is also an evidence-quality and source-consistency problem. Multi-source retrieval may return outdated statements, near-duplicate claims with small but important differences, contradictory web snippets, partially relevant passages, and information whose applicability depends on time, jurisdiction, source authority, or user intent. This situation is common in healthcare, law, public policy, finance, scientific search, and organizational knowledge management. In such settings, a useful information system should not merely produce a fluent answer. It should identify whether the answer is supported, whether sources disagree, and whether a definitive answer is warranted.

Current RAG systems often hide this distinction. Retrieved chunks are usually concatenated into a prompt, and the generator is asked to synthesize an answer. This design implicitly assumes that the retrieved context is reliable, internally coherent, and equally useful. When the assumption fails, an LLM may merge incompatible claims, selectively trust an early-ranked passage, or produce a confident answer even when the retrieved evidence warrants caution \citep{Ji2023HallucinationSurvey,Maynez2020Faithfulness,Lin2022TruthfulQA,Niu2024RAGTruth}. This failure is not fully solved by retrieving more passages. Adding more context can increase the probability of including a correct source, but it can also increase the probability of including mutually incompatible evidence.

Recent work has improved RAG through dense retrieval \citep{Karpukhin2020DPR}, late-interaction retrieval \citep{Khattab2020ColBERT,Santhanam2022ColBERTv2}, reranking \citep{Nogueira2019MonoBERT,Reimers2019SBERT}, self-reflection \citep{Asai2024SelfRAG}, corrective retrieval evaluation \citep{Yan2024CRAG}, and citation-aware generation \citep{Gao2023ALCE}. These advances improve relevance and grounding, but they do not directly quantify epistemic uncertainty caused by contradictions among retrieved sources. A system can retrieve relevant passages and still be unsafe if the passages support incompatible answers. Benchmarks such as CRAG, ConflictBank, and ConflictQA make this issue explicit by showing that RAG systems struggle with dynamic evidence, source disagreement, and conflicts between external evidence and model memory \citep{Yang2024CRAGBenchmark,Su2024ConflictBank,Xie2024ConflictQA}. The core problem is therefore not only whether a document is relevant. The problem is how to represent retrieved information as uncertain evidence and how to propagate that uncertainty to the generator.

This paper introduces \ERAG{}, a framework for conflict-aware, uncertainty-calibrated retrieval-augmented generation. The central design choice is to replace deterministic context injection with probabilistic evidence modeling. Each retrieved chunk is processed by a lightweight evaluator model that extracts candidate factual claims and maps chunk-level support into a Dirichlet evidence representation. This representation follows evidential deep learning, where a Dirichlet distribution encodes both expected support for candidate answers and epistemic uncertainty through total evidence mass \citep{Sensoy2018EDL}. Evidence from multiple chunks is then aggregated using a Dempster-Shafer style fusion rule that explicitly computes conflict mass when sources support mutually exclusive claims \citep{Dempster1967,Shafer1976}. Instead of normalizing conflict away, \ERAG{} transfers unresolved conflict into global uncertainty. At generation time, this uncertainty controls a routing policy that answers directly when evidence is consistent, explains source disagreement when conflict is material, and abstains when the retrieved evidence cannot support a responsible answer.

The contributions of this paper are as follows.
\begin{itemize}[leftmargin=0.55cm]
    \item We formulate multi-source RAG as a probabilistic evidence aggregation problem rather than a deterministic context-concatenation problem, allowing retrieved passages to carry support, ignorance, and contradiction signals.
    \item We introduce a prompt-based evidential extraction module that maps unstructured retrieved text into Dirichlet evidence without fine-tuning the main generator.
    \item We define a conflict-preserving evidence fusion operator that resolves the mass-function inconsistency between Dirichlet evidence, subjective-logic vacuity, and Dempster-Shafer ignorance.
    \item We propose an uncertainty-guided generation policy that routes each query to direct answering, conflict-aware answering, or abstention according to global epistemic uncertainty.
    \item We provide an empirical evaluation on CRAG, ConflictQA, and MuSiQue against Naive RAG, Self-RAG, and Corrective RAG, covering answer quality, hallucination, conflict resolution, calibration, efficiency, and human-audited reliability.
\end{itemize}

The remainder of the paper is organized as follows. Section~\ref{sec:related} reviews related work on RAG, hallucination, uncertainty estimation, and Dempster-Shafer conflict handling. Section~\ref{sec:methodology} presents the \ERAG{} framework, including evidential extraction, claim normalization, conflict-preserving fusion, uncertainty-guided routing, and inference-time call accounting. Section~\ref{sec:experiments} describes the datasets, baselines, implementation settings, evaluation metrics, and human-audit protocol. Section~\ref{sec:results} analyzes the main results, calibration behavior, sensitivity studies, efficiency, ablations, and evaluator reliability. Section~\ref{sec:discussion} discusses implications for expert and intelligent retrieval systems, practical deployment considerations, and limitations. Section~\ref{sec:conclusion} concludes the paper.

\section{Related work}
\label{sec:related}

\subsection{Retrieval-augmented generation}

RAG combines retrieval and generation to improve knowledge-intensive NLP. Early neural approaches such as REALM and RAG retrieved supporting documents to condition generation or language-model pre-training \citep{Guu2020REALM,Lewis2020RAG}. Fusion-in-Decoder demonstrated the value of processing multiple retrieved passages during generation \citep{Izacard2021FiD}, while RETRO showed that retrieval can also support large-scale language modeling \citep{Borgeaud2022RETRO}. These methods motivated a broad ecosystem of retrieval-enhanced foundation models.

The effectiveness of RAG depends heavily on the quality of retrieval. Lexical ranking functions such as BM25 remain important information-retrieval baselines and are often combined with neural retrieval in practical systems \citep{Robertson2009BM25}. Dense Passage Retrieval improved open-domain question answering by learning dual encoders for passage retrieval \citep{Karpukhin2020DPR}. ColBERT and ColBERTv2 introduced late interaction mechanisms that provide a favorable trade-off between accuracy and efficiency \citep{Khattab2020ColBERT,Santhanam2022ColBERTv2}. Cross-encoder rerankers and sentence-transformer models further improve relevance estimation \citep{Nogueira2019MonoBERT,Reimers2019SBERT}. These approaches increase the probability of retrieving useful evidence, but they still treat retrieval quality primarily as relevance to the query. They do not directly represent whether retrieved sources agree with one another.

Recent agentic and self-corrective RAG systems add control after retrieval. Self-RAG learns to retrieve, critique passages, and critique generations through reflection tokens \citep{Asai2024SelfRAG}. Corrective RAG uses a lightweight evaluator to identify poor retrieval and trigger corrective actions \citep{Yan2024CRAG}. Citation-aware methods encourage generated answers to include source attributions \citep{Gao2023ALCE}. These techniques improve reliability, but they are not designed specifically for inter-context conflict. \ERAG{} complements them by making conflict a first-class variable in the retrieval-generation pipeline.

\subsection{Hallucination and factuality in foundation models}

Hallucination has been studied extensively in summarization, question answering, and dialogue \citep{Maynez2020Faithfulness,Ji2023HallucinationSurvey,Manakul2023SelfCheckGPT,Min2023FactScore}. In RAG systems, hallucination can occur even when relevant evidence is available, because the generator may ignore, distort, or overgeneralize from the retrieved context. RAGTruth and related benchmarks show that retrieval augmentation does not eliminate unsupported generation and that context-grounded factuality needs explicit evaluation \citep{Niu2024RAGTruth}. Truthfulness benchmarks also show that models can mimic common falsehoods and express high confidence in incorrect answers \citep{Lin2022TruthfulQA}.

Knowledge conflict introduces a more specific reliability challenge. Conflict can occur between model memory and retrieved evidence, between two retrieved passages, or between heterogeneous sources such as text and knowledge graphs. ConflictQA and related work show that LLMs may be receptive to convincing external evidence, but they may also show confirmation bias when evidence partially agrees with their parametric memory \citep{Xie2024ConflictQA}. ConflictBank focuses on evaluating the influence of knowledge conflicts on LLM behavior \citep{Su2024ConflictBank}. Broader surveys and recent conflict-oriented benchmarks further show that knowledge conflict is becoming a central evaluation problem for LLMs and retrieval-augmented systems \citep{Xu2024KnowledgeConflictsSurvey,Zhao2026ConflictQA,Vergili2024RAGBench}. CRAG stresses RAG systems with dynamic and sometimes ambiguous retrieval scenarios \citep{Yang2024CRAGBenchmark}. These studies motivate methods that distinguish absence of evidence from contradictory evidence and that make uncertainty visible to users.

\subsection{Uncertainty estimation and evidential learning}

Uncertainty estimation is a central issue in trustworthy machine learning. Bayesian approximations, Monte Carlo dropout, deep ensembles, and calibration methods provide different mechanisms for estimating predictive uncertainty \citep{Gal2016Dropout,Lakshminarayanan2017Ensembles,Kendall2017Uncertainties,Ovadia2019TrustUncertainty,Guo2017Calibration}. Expected calibration error and proper scoring rules provide practical tools for measuring whether confidence aligns with empirical accuracy \citep{Gneiting2007ProperScores,Guo2017Calibration}. In LLM-based information systems, calibration is especially important because fluent output can mask weak evidence.

Evidential deep learning (EDL) provides a useful representation for this setting. Instead of producing only a point probability distribution, EDL predicts non-negative evidence parameters that define a Dirichlet distribution over classes \citep{Sensoy2018EDL}. The total evidence controls epistemic uncertainty: low evidence produces high vacuity, while concentrated evidence produces lower uncertainty. This representation is closely connected to subjective logic, where belief masses assigned to singleton outcomes and uncertainty mass assigned to the full frame sum to one \citep{Josang2016SubjectiveLogic}. \ERAG{} uses this mapping to convert retrieved chunks into evidence-bearing observations. The method differs from standard EDL because the evidence vector is produced by a lightweight evaluator model rather than by training a specialized neural classifier for each dataset.

\subsection{Dempster-Shafer evidence theory and conflict handling}

Dempster-Shafer theory represents degrees of belief over sets of hypotheses rather than only over singleton classes \citep{Dempster1967,Shafer1976}. This property makes it attractive for RAG because a retrieved chunk may support a specific answer, provide weak support for several alternatives, or provide only ignorance. Classical Dempster combination normalizes away conflict, which can be problematic when two sources strongly support incompatible singleton claims. Yager's rule and the transferable belief model were introduced partly to address such high-conflict cases by avoiding overconfident normalization \citep{Yager1987,Smets1990}. \ERAG{} uses a conflict-preserving operator that includes Yager's full-conflict transfer as a special case and allows partial conflict transfer through a tunable parameter. This design gives the generator access to both fused support and unresolved disagreement.

\section{Methodology}
\label{sec:methodology}

The goal of \ERAG{} is to make a RAG system aware of disagreement before the final answer is generated. A standard RAG system usually retrieves several passages, places them in the prompt, and relies on the generator to decide what to trust. That approach is simple, but it gives the generator no explicit representation of whether two passages support the same answer, whether one passage is irrelevant, or whether the evidence is too contradictory for a single definitive answer. \ERAG{} adds an intermediate evidential layer between retrieval and generation so that these evidence states become measurable.

The method has four steps. First, retrieved chunks are converted into candidate answer claims and non-negative evidence scores. Second, equivalent claims are normalized into the same candidate so that paraphrases do not create artificial conflict. Third, the candidate-level evidence is fused with a Dempster-Shafer operator that preserves unresolved conflict as epistemic uncertainty. Fourth, the generator receives both the evidence summary and the uncertainty score, and it is routed to direct answering, conflict-aware answering, or abstention. The mathematical notation below formalizes these steps, while the surrounding explanations indicate what each quantity means operationally.

\subsection{Problem setting}

Let $q$ denote a user query and let $C_q=\{c_1,c_2,\ldots,c_k\}$ denote the top-$k$ retrieved chunks returned by the retriever and reranker. In ordinary RAG, the generator conditions on $q$ and a concatenation of $C_q$. In \ERAG{}, the same retrieved chunks are treated as observations that may provide support, ignorance, or contradiction. This formulation exposes three variables that are hidden by plain concatenation: which candidate answer each chunk supports, how much evidence each chunk provides, and whether chunks conflict.

We define a finite candidate-claim set $A_q=\{a_1,a_2,\ldots,a_M\}$ induced by the retrieved chunks. A claim may be an answer string, entity mention, date, numerical value, or short proposition. The Dempster-Shafer frame of discernment is
\begin{equation}
    \Frame_q=A_q.
    \label{eq:frame}
\end{equation}
The frame contains only mutually distinguishable candidate claims. Ignorance is not represented as an additional singleton class. Instead, ignorance is represented as mass assigned to the full frame $\Frame_q$. This distinction is essential. A retrieved chunk that is irrelevant or insufficient does not provide evidence for a special ``unknown'' answer. It provides little or no singleton evidence, which produces high vacuity.

For each chunk $c_i$, the evaluator outputs a non-negative evidence vector $e_i\in\R_{\geq0}^{M}$. The corresponding Dirichlet parameter vector is
\begin{equation}
    \alpha_i=e_i+\mathbf{1}, \qquad S_i=\sum_{j=1}^{M}\alpha_{ij},
    \label{eq:dirichlet}
\end{equation}
where $S_i$ is the total Dirichlet concentration. The expected probability of claim $a_j$ is
\begin{equation}
    \hat{p}_{ij}=\frac{\alpha_{ij}}{S_i}.
    \label{eq:expected_prob}
\end{equation}
Following the subjective-logic mapping used in EDL, singleton belief and vacuity are defined as
\begin{equation}
    b_{ij}=\frac{e_{ij}}{S_i}, \qquad u_i=\frac{M}{S_i}.
    \label{eq:belief_vacuity}
\end{equation}
This yields a valid mass function because
\begin{equation}
    \sum_{j=1}^{M}b_{ij}+u_i=\frac{\sum_{j=1}^{M}e_{ij}+M}{S_i}=1.
    \label{eq:mass_validity}
\end{equation}
The induced Dempster-Shafer mass function is therefore
\begin{equation}
    m_i(\{a_j\})=b_{ij}, \qquad m_i(\Frame_q)=u_i.
    \label{eq:mass_function}
\end{equation}
This equation is the bridge between EDL and Dempster-Shafer reasoning. The singleton mass $m_i(\{a_j\})$ means that chunk $c_i$ specifically supports claim $a_j$. The frame mass $m_i(\Frame_q)$ means that the chunk does not provide enough evidence to distinguish among candidate claims. No mass is assigned to an additional unknown singleton. If a chunk is irrelevant, the evaluator sets $e_{ij}\approx0$ for all $j$, yielding $u_i\approx1$. If a chunk strongly supports one claim, it assigns high evidence to that singleton, yielding low vacuity. This design prevents irrelevant context from being interpreted as positive evidence for a fabricated ``unknown'' answer category.

\subsection{Evidential claim extraction}

The evidential evaluator receives the query, the retrieved chunk, and the current claim inventory. It returns a structured object containing the extracted claim, normalized answer value, stance, evidence rationale, and evidence scores. The evaluator used in the reported experiments is Llama-3-8B-Instruct. It is intentionally smaller than the Llama-3-70B-Instruct generator so that uncertainty estimation does not require multiple full-generator passes.

The evaluator performs three functions. First, it extracts the answer-bearing proposition in the chunk, if one exists. Second, it normalizes surface forms such as dates, numbers, entity aliases, and paraphrases. Third, it assigns non-negative evidence to candidate claims. If the chunk provides no answer-bearing information, the evidence vector is near zero and the chunk contributes ignorance. If the chunk supports a claim that contradicts another retrieved claim, it assigns evidence to the competing singleton rather than to an unknown bucket.

The prompt-based design avoids training a new evidential model for every domain. It also keeps the architecture modular. The evaluator can be replaced by a fine-tuned verifier, a natural-language inference model, a domain-specific classifier, or a stronger LLM-as-evaluator. The downstream fusion module requires only the normalized evidence vectors and does not depend on the evaluator architecture.

\subsection{Candidate claim normalization and alignment}

The word ``alignment'' here refers to aligning answer claims across retrieved chunks, not aligning the paper's claims. It is necessary because open-domain answers are not fixed labels. Different passages can express the same fact with different surface forms, while other passages may express genuinely incompatible facts. Without this step, a system could incorrectly treat ``New York City'', ``NYC'', and ``New York, NY'' as three conflicting answers, or it could incorrectly merge two different dates that should remain separate.

We use a two-stage procedure. First, lexical and semantic normalization merges equivalent surface forms, including abbreviations, aliases, date formats, numerical formatting, and minor paraphrases. Second, incompatible claims remain distinct elements in $A_q$. Numerical and temporal claims are treated as incompatible when their normalized values differ beyond a task-specific tolerance. Entity claims are treated as incompatible when canonical identifiers differ. Proposition claims are treated as incompatible when semantic equivalence cannot be established and the evaluator marks the stance as competing.

This stage prevents paraphrase diversity from being mistaken for evidence conflict. Two chunks that state the same answer differently should increase evidence for the same claim. Two chunks that state different answers should increase conflict. The procedure is conservative: uncertain merges are avoided, and downstream uncertainty is allowed to increase rather than forcing premature agreement.

\subsection{Conflict-preserving fusion}

Let $m_a$ and $m_b$ be two mass functions over singleton claims and the full frame. The pairwise conflict mass is
\begin{equation}
    K(m_a,m_b)=\sum_{r\neq s}m_a(\{a_r\})m_b(\{a_s\}).
    \label{eq:conflict_mass}
\end{equation}
For a singleton claim $a_j$, the non-conflicting conjunctive support is
\begin{equation}
    n_j=m_a(\{a_j\})m_b(\{a_j\}) + m_a(\{a_j\})m_b(\Frame_q) + m_a(\Frame_q)m_b(\{a_j\}).
    \label{eq:nonconf_support}
\end{equation}
The joint ignorance term is
\begin{equation}
    h=m_a(\Frame_q)m_b(\Frame_q).
    \label{eq:joint_ignorance}
\end{equation}
The conjunctive components are disjoint and exhaustive over the frame, so $h+K+\sum_{j=1}^{M}n_j=1$. This identity shows that the transfer step operates on a valid evidence decomposition rather than introducing an unaccounted mass term. Classical normalized Dempster combination divides non-empty intersections by $1-K$. Under severe conflict, this normalization can produce overconfident singleton masses. \ERAG{} instead uses a conflict-transfer parameter $\lambda\in[0,1]$ and assigns
\begin{equation}
    m_{ab}(\Frame_q)=h+\lambda K.
    \label{eq:frame_transfer}
\end{equation}
The remaining mass is assigned to singleton claims by scaling their non-conflicting support:
\begin{equation}
    m_{ab}(\{a_j\})=
    \begin{cases}
    \dfrac{(1-h-\lambda K)n_j}{\sum_{\ell=1}^{M} n_\ell}, & \sum_{\ell=1}^{M}n_\ell>0,\\[1.2em]
    0, & \sum_{\ell=1}^{M}n_\ell=0.
    \end{cases}
    \label{eq:singleton_transfer}
\end{equation}
The operator is valid because singleton masses and frame mass sum to one whenever $0\leq h+\lambda K\leq1$. The condition is satisfied since $h$ and $K$ are disjoint components of the conjunctive mass decomposition. When $\lambda=1$, all conflict is transferred to ignorance, matching the intuition of Yager-style high-conflict handling. When $\lambda=0$, conflict is not transferred to the frame and remaining mass is redistributed among non-conflicting singleton supports. Intermediate values allow partial transfer and can be selected on validation data.

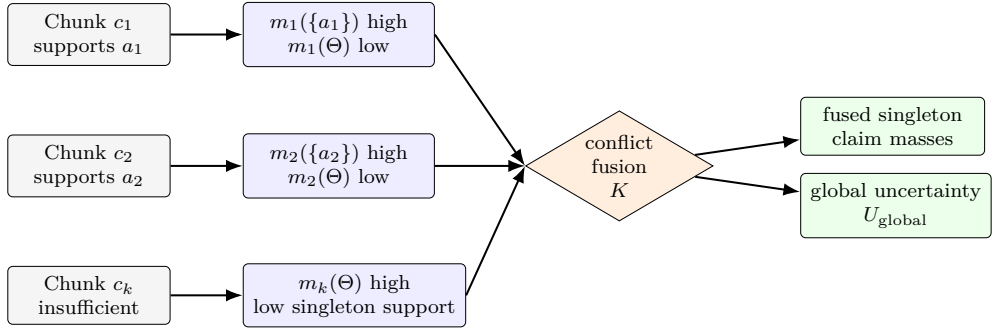
\begin{figure}[H]
\centering
\resizebox{0.95\linewidth}{!}{%
\begin{tikzpicture}[
    node distance=0.95cm and 1.0cm,
    chunk/.style={draw, rounded corners=2pt, minimum width=2.25cm, minimum height=0.8cm, align=center, font=\scriptsize, fill=gray!8},
    mass/.style={draw, rounded corners=2pt, minimum width=2.65cm, minimum height=0.8cm, align=center, font=\scriptsize, fill=blue!7},
    fuse/.style={draw, diamond, aspect=1.7, align=center, font=\scriptsize, fill=orange!13, inner sep=1pt},
    outbox/.style={draw, rounded corners=2pt, minimum width=2.55cm, minimum height=0.8cm, align=center, font=\scriptsize, fill=green!8},
    arrow/.style={-Latex, thick}
]
\node[chunk] (c1) {Chunk $c_1$\\supports $a_1$};
\node[chunk, below=of c1] (c2) {Chunk $c_2$\\supports $a_2$};
\node[chunk, below=of c2] (ck) {Chunk $c_k$\\insufficient};
\node[mass, right=of c1] (m1) {$m_1(\{a_1\})$ high\\$m_1(\Theta)$ low};
\node[mass, right=of c2] (m2) {$m_2(\{a_2\})$ high\\$m_2(\Theta)$ low};
\node[mass, right=of ck] (mk) {$m_k(\Theta)$ high\\low singleton support};
\node[fuse, right=1.25cm of m2] (fus) {conflict\\fusion\\$K$};
\node[outbox, right=1.2cm of fus, yshift=0.55cm] (claims) {fused singleton\\claim masses};
\node[outbox, right=1.2cm of fus, yshift=-0.55cm] (unc) {global uncertainty\\$U_{\mathrm{global}}$};
\draw[arrow] (c1) -- (m1);
\draw[arrow] (c2) -- (m2);
\draw[arrow] (ck) -- (mk);
\draw[arrow] (m1.east) -- (fus.west);
\draw[arrow] (m2.east) -- (fus.west);
\draw[arrow] (mk.east) -- (fus.west);
\draw[arrow] (fus) -- (claims);
\draw[arrow] (fus) -- (unc);
\end{tikzpicture}}
\caption{Conflict-preserving fusion of retrieved evidence. Equivalent claims reinforce the same singleton mass, incompatible claims contribute to conflict mass $K$, and unresolved conflict is transferred into ignorance over the frame rather than normalized into overconfident support. The pairwise conflict $K$ is retained as a diagnostic statistic and is not added directly to $U_{\mathrm{global}}$.}
\label{fig:fusion_visual}
\end{figure}

For $k$ retrieved chunks, the pairwise operator $\otimes_{\lambda}$ is applied iteratively, where $\otimes_{\lambda}$ denotes the conflict-preserving fusion operator defined in the equations above:
\begin{equation}
    m^{(1)}=m_1, \qquad m^{(t)}=m^{(t-1)}\otimes_{\lambda} m_t, \quad t=2,\ldots,k.
    \label{eq:iter_fusion}
\end{equation}
The global uncertainty score used for routing is the final mass assigned to the full frame:
\begin{equation}
    U_{\mathrm{global}}=m^{(k)}(\Frame_q).
    \label{eq:global_uncertainty}
\end{equation}
The average unfused pairwise conflict
\begin{equation}
    \bar{K}=\frac{2}{k(k-1)}\sum_{1\leq r<s\leq k}K(m_r,m_s)
    \label{eq:pairwise_conflict_diagnostic}
\end{equation}
is retained only as a diagnostic statistic. It is not added again to $U_{\mathrm{global}}$, which avoids double-counting conflict after the transfer step. Figure~\ref{fig:fusion_visual} illustrates how singleton support, conflict mass, and ignorance are separated during this fusion process.

\subsection{Uncertainty-guided generation}

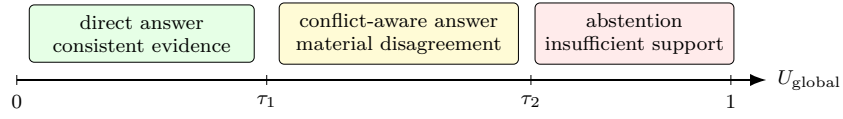
\begin{figure}[H]
\centering
\resizebox{0.82\linewidth}{!}{%
\begin{tikzpicture}[
    band/.style={draw, rounded corners=2pt, minimum height=0.75cm, align=center, font=\scriptsize},
    arrow/.style={-Latex, thick}
]
\draw[arrow] (0,0) -- (10.5,0) node[right, font=\scriptsize]{$U_{\mathrm{global}}$};
\draw (0,0.07) -- (0,-0.07) node[below, font=\scriptsize]{0};
\draw (3.5,0.07) -- (3.5,-0.07) node[below, font=\scriptsize]{$\tau_1$};
\draw (7.2,0.07) -- (7.2,-0.07) node[below, font=\scriptsize]{$\tau_2$};
\draw (10,0.07) -- (10,-0.07) node[below, font=\scriptsize]{1};
\node[band, fill=green!10, minimum width=3.15cm] at (1.75,0.65) {direct answer\\consistent evidence};
\node[band, fill=yellow!20, minimum width=3.35cm] at (5.35,0.65) {conflict-aware answer\\material disagreement};
\node[band, fill=red!8, minimum width=2.55cm] at (8.65,0.65) {abstention\\insufficient support};
\end{tikzpicture}}
\caption{Uncertainty-guided generation routes. The thresholds separate ordinary evidence-supported answering from conflict explanation and abstention.}
\label{fig:routing_visual}
\end{figure}

The final generator receives the query, the ranked evidence summary, the fused mass distribution, and $U_{\mathrm{global}}$. The routing policy is governed by two thresholds $\tau_1$ and $\tau_2$, where $0\leq\tau_1<\tau_2\leq1$:
\begin{equation}
\text{route}(q)=
\begin{cases}
\text{direct answer}, & U_{\mathrm{global}}<\tau_1,\\
\text{conflict-aware answer}, & \tau_1\leq U_{\mathrm{global}}<\tau_2,\\
\text{abstention}, & U_{\mathrm{global}}\geq\tau_2.
\end{cases}
    \label{eq:routing_policy}
\end{equation}
In direct-answer mode, the generator answers using the highest-mass claim and cites supporting chunks. In conflict-aware mode, the generator explicitly states that retrieved sources disagree, summarizes competing claims, and avoids presenting a single answer as settled. In abstention mode, the generator declines to provide a definitive answer and explains what additional evidence would be needed. The three route regions induced by the uncertainty thresholds are visualized in Figure~\ref{fig:routing_visual}.

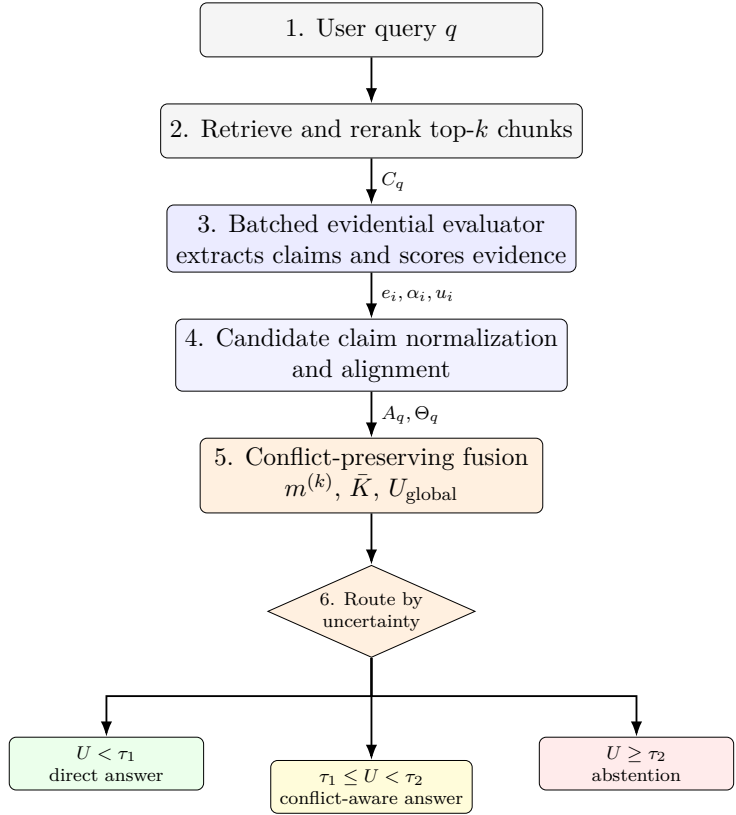
\begin{figure}[H]
\centering
\resizebox{0.70\linewidth}{!}{%
\begin{tikzpicture}[
    node distance=0.70cm,
    stage/.style={draw, rounded corners=3pt, align=center, minimum height=0.82cm, minimum width=5.2cm, font=\small},
    decision/.style={draw, diamond, aspect=2.25, align=center, inner sep=1.2pt, font=\scriptsize, fill=orange!12},
    output/.style={draw, rounded corners=3pt, align=center, minimum height=0.72cm, minimum width=2.95cm, font=\scriptsize},
    arrow/.style={-Latex, thick}
]
\node[stage, fill=gray!8] (q) {1. User query $q$};
\node[stage, fill=gray!8, below=of q] (retr) {2. Retrieve and rerank top-$k$ chunks};
\node[stage, fill=blue!8, below=of retr] (eval) {3. Batched evidential evaluator\\extracts claims and scores evidence};
\node[stage, fill=blue!5, below=of eval] (align) {4. Candidate claim normalization\\and alignment};
\node[stage, fill=orange!12, below=of align] (fusion) {5. Conflict-preserving fusion\\$m^{(k)}$, $\bar{K}$, $U_{\mathrm{global}}$};
\node[decision, below=0.78cm of fusion] (route) {6. Route by\\uncertainty};
\node[output, fill=green!8, below left=1.55cm and 1.75cm of route] (direct) {$U<\tau_1$\\direct answer};
\node[output, fill=yellow!18, below=1.55cm of route] (aware) {$\tau_1\leq U<\tau_2$\\conflict-aware answer};
\node[output, fill=red!8, below right=1.55cm and 1.75cm of route] (abs) {$U\geq\tau_2$\\abstention};
\coordinate (split) at ($(route.south)+(0,-0.58)$);
\draw[arrow] (q) -- (retr);
\draw[arrow] (retr) -- node[right, font=\scriptsize]{$C_q$} (eval);
\draw[arrow] (eval) -- node[right, font=\scriptsize]{$e_i,\alpha_i,u_i$} (align);
\draw[arrow] (align) -- node[right, font=\scriptsize]{$A_q,\Theta_q$} (fusion);
\draw[arrow] (fusion) -- (route);
\draw[arrow] (route.south) -- (split) -| (direct.north);
\draw[arrow] (route.south) -- (split) -- (aware.north);
\draw[arrow] (route.south) -- (split) -| (abs.north);
\end{tikzpicture}}
\caption{Overview of \ERAG{}. The sequence starts with query-specific retrieval, converts chunks into evidential masses, fuses source evidence while preserving conflict, and routes the generator according to global uncertainty.}
\label{fig:overview}
\end{figure}

The complete end-to-end flow of \ERAG{} is shown in Figure~\ref{fig:overview}. The diagram emphasizes that retrieval, evidence extraction, fusion, and generation routing are sequential stages rather than independent modules.

\subsection{Algorithm and LLM-call accounting}

The boxed inference procedure summarizes inference. The evaluator is conceptually applied to each retrieved chunk, but implementation can batch these chunk evaluations into a single evaluator forward pass. This distinction is important for latency accounting. With top-$k$ retrieval, the logical number of evaluator decisions is $k$, and the logical number of LLM decisions is $k+1$ including final generation. In the reported implementation, those $k$ evaluator decisions are issued as one batched evaluator pass, followed by one generator pass. The efficiency table therefore separates batched LLM passes from per-chunk logical decisions. Figure~\ref{fig:algorithm} gives the full inference procedure and links each stage to the corresponding mathematical definition.

\begin{figure}[H]
\centering
\fbox{%
\begin{minipage}{0.94\linewidth}
\textbf{\ERAG{} inference procedure for one query.}
\begin{enumerate}[leftmargin=0.55cm]
\item \textbf{Input:} query $q$, retriever $R$, reranker $G_r$, top-$k$, evaluator $E$, generator $G$, thresholds $\tau_1,\tau_2$, transfer parameter $\lambda$.
\item Retrieve and rerank $C_q=\{c_1,\ldots,c_k\}$ using $R$ and $G_r$.
\item Batch $\{(q,c_i)\}_{i=1}^{k}$ through evaluator $E$ to obtain normalized claims and evidence vectors $e_i$.
\item Align claims into $A_q=\{a_1,\ldots,a_M\}$ and construct the frame $\Frame_q=A_q$.
\item Convert each $e_i$ into $\alpha_i$, $b_i$, $u_i$, and mass function $m_i$ using Eqs.~\eqref{eq:dirichlet}--\eqref{eq:mass_function}.
\item Fuse $m_1,\ldots,m_k$ iteratively with the conflict-preserving operator in Eqs.~\eqref{eq:conflict_mass}--\eqref{eq:global_uncertainty}.
\item Set $U_{\mathrm{global}}=m^{(k)}(\Frame_q)$ using Eq.~\eqref{eq:global_uncertainty}.
\item Route generator $G$ using Eq.~\eqref{eq:routing_policy}.
\item \textbf{Output:} generated response, selected route, fused evidence, and uncertainty metadata.
\end{enumerate}
\end{minipage}}
\caption{Inference procedure for \ERAG{}. The procedure distinguishes per-chunk evidence decisions from batched evaluator execution.}
\label{fig:algorithm}
\end{figure}

\section{Experimental setup}
\label{sec:experiments}

\subsection{Datasets}

We evaluate \ERAG{} on three benchmark families selected to separate three capabilities: ordinary factual answering, explicit knowledge-conflict handling, and non-conflict multi-hop synthesis. The broader experimental design follows the tradition of open-domain and reading-comprehension evaluation established by Natural Questions and SQuAD, while focusing on the additional difficulty of retrieval-grounded conflict \citep{Kwiatkowski2019NQ,Rajpurkar2016SQuAD}. CRAG measures factual RAG behavior across diverse domains, question types, entity popularity levels, and temporal dynamics \citep{Yang2024CRAGBenchmark}. The ambiguous and incorrect-context portions are used to stress-test conflict behavior. ConflictQA evaluates the behavior of LLMs when external evidence conflicts with parametric memory or when the provided evidence induces incompatible answer commitments \citep{Xie2024ConflictQA}. MuSiQue provides a non-conflict multi-hop setting, allowing us to test whether uncertainty routing harms ordinary compositional reasoning \citep{Trivedi2022MuSiQue}. FEVER and HotpotQA are not primary test sets in this study; they are used as methodological precedents for evidence-based factuality and multi-hop evaluation \citep{Thorne2018FEVER,Yang2018HotpotQA}. Table~\ref{tab:datasets} summarizes the scale, conflict setting, role, and metrics for each primary benchmark.

\begin{table}[H]
\centering
\caption{Dataset characteristics for the evaluation benchmarks. The table reports the benchmark-level properties that determine how each dataset stresses retrieval, conflict modeling, and generation.}
\label{tab:datasets}
\scriptsize
\resizebox{\linewidth}{!}{%
\begin{tabular}{p{0.12\textwidth}p{0.17\textwidth}p{0.18\textwidth}p{0.20\textwidth}p{0.21\textwidth}p{0.18\textwidth}}
\toprule
Dataset & Scale and source & Domain and question coverage & Evidence or conflict setting & Use in this paper & Primary metrics \\
\midrule
CRAG & 4,409 factual QA pairs with associated web and KG-style retrieval interfaces & Five domains and eight question categories, with entities ranging from popular to long-tail and temporal dynamics ranging from stable facts to rapidly changing facts & Includes standard cases and cases where retrieved context can be ambiguous, incomplete, stale, or incorrect & Main benchmark for standard QA and the CRAG ambiguous subset used for conflict-stress evaluation & EM, F1, ROUGE-L, BERTScore, hallucination rate, CRR, ECE, abstention \\
ConflictQA & Controlled knowledge-conflict benchmark derived from parametric-memory elicitation and counter-memory construction; the conflict-specific evaluation uses 1,000 processed instances marked by the benchmark construction as memory-versus-evidence conflicts & Open-domain factual questions designed to expose whether models follow external evidence or prior parametric knowledge & External evidence may be consistent with, partially consistent with, or contradictory to model memory & Used to evaluate source disagreement, memory-versus-retrieval conflict, and conflict-aware non-definitive answering & EM, F1, hallucination rate, CRR, refusal precision, refusal recall, refusal F1, ECE \\
MuSiQue & Approximately 25K 2-hop to 4-hop multi-hop QA examples constructed through single-hop question composition & Multi-hop questions requiring connected reasoning over evidence passages & Not designed as a contradiction benchmark; evidence is used to test compositional synthesis under ordinary retrieval conditions & Used as a non-conflict stress test to verify that evidential routing does not unnecessarily suppress valid multi-hop answers & EM, F1, ROUGE-L, BERTScore, abstention rate \\
\bottomrule
\end{tabular}}

\end{table}

The same preprocessing principles are applied across datasets. Questions, gold answers, retrieved chunks, and model responses are stored with stable identifiers. For conflict-focused subsets, each instance is additionally labeled with the expected conflict condition, the candidate answer claims, and whether a responsible response should answer directly, explain disagreement, or abstain. This metadata supports both the automated metrics and the stratified human audit described in Section~\ref{sec:human_eval}.

\subsection{Baselines}

We compare \ERAG{} with three baselines. Naive RAG retrieves top-$k$ chunks and inserts them directly into the generator prompt. Self-RAG adds reflection tokens to decide when to retrieve, critique passages, and critique its own generation \citep{Asai2024SelfRAG}. Corrective RAG adds a retrieval evaluator and triggers corrective actions when retrieval appears unreliable \citep{Yan2024CRAG}. All baselines are run under a shared evaluation harness with the same retrieval corpus, generator, maximum context budget, and maximum generation length. This controlled setting isolates the contribution of the evidential conflict module. It may differ from the original published hyperparameter settings of Self-RAG and Corrective RAG, so the results should be interpreted as matched-system comparisons rather than direct reproduction of the original papers.

\subsection{Implementation details}

The reported implementation uses top-$k$ retrieval with $k=5$. Retrieved chunks are reranked before evidential evaluation. The evidential evaluator is Llama-3-8B-Instruct, and the final generator is Llama-3-70B-Instruct \citep{Grattafiori2024Llama3}. The framework is not tied to this evaluator family; comparable open instruction-tuned models, including Mistral-style models, can be used when their structured-output behavior is validated \citep{Jiang2023Mistral7B}. Retrieval uses bge-large-en-v1.5, and reranking uses bge-reranker-v2-m3 from the BGE embedding family \citep{Chen2024BGEM3}. Inference is served with vLLM v0.4.0 to reduce latency under batched evaluation \citep{Kwon2023vLLM}. Model and tokenizer management use the Transformers ecosystem \citep{Wolf2020Transformers}. The default transfer parameter is $\lambda=0.6$. The routing thresholds are $\tau_1=0.35$ and $\tau_2=0.65$. The transfer parameter and routing thresholds were selected before final testing on validation data only. We used the released development partitions when the benchmark provided them; for conflict subsets without a separate development partition, we created a stratified 10\% development split from the training portion using the same conflict labels used for evaluation. No instance from the final evaluation partitions used in the reported tables was used for threshold or $\lambda$ selection. MuSiQue was not used for tuning and served as a post-selection non-conflict sanity check. Expected calibration error uses $B=15$ equal-width confidence bins. All main reported results are averaged over three random seeds, $\{42,123,999\}$, with standard deviations shown in the tables. Paired bootstrap tests are used for significance testing against Corrective RAG on the same evaluated instances \citep{Efron1979Bootstrap}.

The evaluator prompt is constrained to return valid JSON. Invalid JSON responses are repaired using a conservative schema validator. If repair fails, the chunk is assigned near-zero singleton evidence, producing high uncertainty rather than hidden confidence. This conservative fallback is important because the framework should not convert evaluator formatting errors into apparent factual support.

\subsection{Evaluation metrics}
\label{sec:evaluation_metrics}

We evaluate answer quality, reliability under conflict, calibration, refusal behavior, and computational efficiency. Let $n$ denote the number of evaluated instances, $y_i$ the normalized gold answer for instance $i$, $\hat{y}_i$ the normalized model answer, and $r_i$ the generated response text.

\textbf{Exact Match.} EM measures the percentage of predictions that exactly match a gold answer after normalization:
\begin{equation}
    \mathrm{EM}=\frac{1}{n}\sum_{i=1}^{n}\mathbb{I}\left[\mathrm{norm}(\hat{y}_i)=\mathrm{norm}(y_i)\right].
\end{equation}
This metric is strict and is most informative for short factual answers such as dates, names, and entities.

\textbf{Token-level F1.} Token-level F1 measures partial lexical overlap between the prediction and reference. For each instance, precision and recall are
\begin{equation}
    P_i=\frac{|T(\hat{y}_i)\cap T(y_i)|}{|T(\hat{y}_i)|}, \qquad
    R_i=\frac{|T(\hat{y}_i)\cap T(y_i)|}{|T(y_i)|},
\end{equation}
where $T(\cdot)$ is the normalized token set or multiset. The instance-level F1 and corpus-level F1 are
\begin{equation}
    F1_i=\frac{2P_iR_i}{P_i+R_i}, \qquad
    \mathrm{F1}=\frac{1}{n}\sum_{i=1}^{n}F1_i.
\end{equation}
If both precision and recall are zero, $F1_i$ is defined as zero.

\textbf{ROUGE-L.} ROUGE-L captures sequence-level overlap using the longest common subsequence (LCS) between prediction and reference \citep{Lin2004ROUGE}. Let $L_i$ be the LCS length, $m_i$ the reference length, and $\ell_i$ the prediction length. Then
\begin{equation}
    R^{L}_i=\frac{L_i}{m_i}, \qquad P^{L}_i=\frac{L_i}{\ell_i}, \qquad
    \mathrm{ROUGE\mbox{-}L}_i=\frac{(1+\beta^2)R^{L}_iP^{L}_i}{R^{L}_i+\beta^2P^{L}_i}.
\end{equation}
We use the standard F-measure form and average over all instances.

\textbf{BERTScore.} BERTScore estimates semantic similarity using contextual token embeddings \citep{Zhang2020BERTScore}. If $x_j$ and $z_k$ are contextual embeddings of prediction and reference tokens, the precision and recall components are
\begin{equation}
    P^{B}_i=\frac{1}{|\hat{y}_i|}\sum_{x_j\in \hat{y}_i}\max_{z_k\in y_i} \cos(x_j,z_k), \qquad
    R^{B}_i=\frac{1}{|y_i|}\sum_{z_k\in y_i}\max_{x_j\in \hat{y}_i} \cos(x_j,z_k),
\end{equation}
with the final BERTScore F1 computed as $2P^{B}_iR^{B}_i/(P^{B}_i+R^{B}_i)$.

\textbf{Hallucination rate.} Hallucination is evaluated at the response level. Let $H_i=1$ if response $r_i$ contains at least one factual assertion that is unsupported by, or contradicted by, the retrieved evidence and gold annotation; otherwise $H_i=0$. Hallucination rate is
\begin{equation}
    \mathrm{Hallucination\ Rate}=\frac{1}{n}\sum_{i=1}^{n}H_i.
\end{equation}
Lower values indicate fewer unsupported factual commitments.

\textbf{Conflict Resolution Rate.} CRR is computed only over the subset $\mathcal{C}$ of genuinely conflicting instances. Let $C_i=1$ if the response explicitly recognizes source disagreement, summarizes the competing claims without collapsing them into an unsupported single answer, and avoids overconfident commitment. Then
\begin{equation}
    \mathrm{CRR}=\frac{1}{|\mathcal{C}|}\sum_{i\in\mathcal{C}}C_i.
\end{equation}
This metric rewards conflict-aware information behavior rather than blind abstention.

\textbf{Refusal precision, recall, and F1.} Abstention is defined operationally at the output level for all systems, including Naive RAG. A response is counted as an abstention when it explicitly states inability to answer, insufficient evidence, source disagreement that prevents a definitive answer, or an equivalent non-definitive stance. This means Naive RAG can have a nonzero abstention rate when the generator spontaneously produces phrases such as ``I do not know'' or ``the provided context is insufficient'', even though it has no explicit abstention controller. Let $R_i=1$ denote a refusal or conflict-aware non-definitive response. Let $U_i=1$ denote that the instance is genuinely unanswerable from the retrieved evidence or contains unresolved conflict. Refusal precision and recall are
\begin{equation}
    \mathrm{Refusal\ Precision}=\frac{\sum_i \mathbb{I}[R_i=1\land U_i=1]}{\sum_i \mathbb{I}[R_i=1]},
\end{equation}
\begin{equation}
    \mathrm{Refusal\ Recall}=\frac{\sum_i \mathbb{I}[R_i=1\land U_i=1]}{\sum_i \mathbb{I}[U_i=1]}.
\end{equation}
The refusal F1 score is
\begin{equation}
    \mathrm{Refusal\ F1}=\frac{2\cdot \mathrm{Precision}\cdot \mathrm{Recall}}{\mathrm{Precision}+\mathrm{Recall}}.
\end{equation}
A high refusal score requires the model to abstain when appropriate without refusing answerable cases unnecessarily.

\textbf{Expected Calibration Error.} ECE measures the mismatch between confidence and empirical correctness \citep{Guo2017Calibration}. Responses are partitioned into $B=15$ equal-width confidence bins $\mathcal{B}_1,\ldots,\mathcal{B}_B$. We compute
\begin{equation}
    \mathrm{ECE}=\sum_{b=1}^{B}\frac{|\mathcal{B}_b|}{n}\left|\mathrm{acc}(\mathcal{B}_b)-\mathrm{conf}(\mathcal{B}_b)\right|.
\end{equation}
For \ERAG{}, confidence is the fused singleton mass assigned to the selected answer:
\begin{equation}
    \mathrm{conf}_i=\max_{a_j\in A_q}m^{(k)}_i(\{a_j\}).
\end{equation}
We do not multiply this value again by $(1-U_{\mathrm{global}})$ because the fused mass function already allocates a portion of total mass to the full frame. A second uncertainty multiplier would double-discount uncertainty and would make the confidence statistic less interpretable for calibration.

\textbf{Efficiency.} Latency is the mean wall-clock inference time per query. Relative inference cost is normalized by Naive RAG:
\begin{equation}
    \mathrm{Relative\ Cost}(M)=\frac{\mathrm{Latency}(M)}{\mathrm{Latency}(\mathrm{Naive\ RAG})}.
\end{equation}
We report both logical LLM decisions and batched model passes because \ERAG{} evaluates $k$ chunks but executes those evaluator decisions in one batched pass.

\subsection{LLM judge and human audit}
\label{sec:human_eval}

Hallucination rate and CRR require judgments about factual support, contradiction, and whether a refusal is appropriate. We therefore combine automated judging with a stratified human audit. The LLM judge receives the question, gold answer when available, retrieved evidence, candidate claims, route decision, and generated response. It assigns binary labels for unsupported factual assertion, contradicted factual assertion, conflict recognition, appropriate refusal, and answer support. The judge is instructed to evaluate only the visible response against the provided evidence and not to reward fluent unsupported speculation.

We conducted a stratified human audit of 200 responses sampled across models, datasets, routing decisions, and correctness labels. The sample was balanced to include direct answers, conflict-aware answers, abstentions, correct outputs, and incorrect outputs from all evaluated systems. Two annotators independently labeled each response along four dimensions: factual support, contradiction, conflict recognition, and refusal appropriateness. Factual support indicates whether the central answer is entailed by the retrieved evidence. Contradiction indicates whether the response asserts a claim that conflicts with at least one reliable retrieved source or the gold annotation. Conflict recognition indicates whether the response explicitly acknowledges incompatible evidence when such evidence is present. Refusal appropriateness indicates whether abstention or a non-definitive answer is justified by the evidence state.

Inter-annotator agreement was measured using Cohen's $\kappa$ for pairwise agreement between the two initial annotators before adjudication \citep{Cohen1960Kappa}, and disagreements were resolved through adjudication. During adjudication, annotators reviewed the evidence passages and response together; if disagreement remained, a third annotator made the final label. The adjudicated labels were used to audit the LLM judge and to verify that the metric definitions above were applied consistently. This human evaluation is particularly important for conflict cases because a response can be lexically similar to a gold answer while still being unreliable if it ignores source disagreement. Table~\ref{tab:human_audit_dimensions} defines the annotation dimensions used for this audit.

\begin{table}[H]
\centering
\caption{Human-audit annotation scheme. Each sampled response is independently labeled by two annotators before adjudication.}
\label{tab:human_audit_dimensions}
\small
\begin{tabularx}{\linewidth}{>{\raggedright\arraybackslash}p{0.24\linewidth} >{\raggedright\arraybackslash}p{0.50\linewidth} >{\raggedright\arraybackslash}p{0.18\linewidth}}
\toprule
Dimension & Positive label definition & Used for \\
\midrule
Factual support & The response's central factual claim is supported by the retrieved evidence or gold annotation. & EM, F1 audit, factuality \\
Contradiction & The response asserts at least one claim contradicted by retrieved evidence or gold annotation. & Hallucination rate \\
Conflict recognition & The response explicitly states or clearly explains that sources disagree. & CRR \\
Refusal appropriateness & The response refuses, hedges, or gives a non-definitive answer only when evidence is insufficient or conflicting. & Refusal precision and recall \\
\bottomrule
\end{tabularx}
\end{table}

\section{Results}
\label{sec:results}

\subsection{Overall question-answering performance}

All headline experiments were re-executed under the revised evidential formalism, using $\Theta=A_q$ and the $\lambda$-transfer fusion operator. The standard benchmark results in Table~\ref{tab:overall} indicate that \ERAG{} remains competitive on ordinary question-answering quality while adding uncertainty-aware conflict handling. Corrective RAG remains slightly stronger on CRAG standard EM and F1, whereas \ERAG{} obtains the best ConflictQA standard F1 among the matched systems.

\begin{table}[H]
\centering
\caption{Question-answering performance on standard benchmark subsets. Values are mean $\pm$ standard deviation over three seeds.}
\label{tab:overall}
\small
\begin{tabularx}{\linewidth}{>{\raggedright\arraybackslash}p{0.27\linewidth}YYYY}
\toprule
Model & \makecell{CRAG\\EM} & \makecell{CRAG\\F1} & \makecell{CQA\\EM} & \makecell{CQA\\F1} \\
\midrule
Naive RAG      & 45.2 $\pm$ 0.5 & 52.1 $\pm$ 0.4 & 38.4 $\pm$ 0.6 & 44.2 $\pm$ 0.5 \\
Self-RAG       & 51.4 $\pm$ 0.4 & 58.3 $\pm$ 0.5 & 45.1 $\pm$ 0.4 & 51.3 $\pm$ 0.6 \\
Corrective RAG & \textbf{54.1 $\pm$ 0.3} & \textbf{61.2 $\pm$ 0.4} & 48.6 $\pm$ 0.5 & 55.4 $\pm$ 0.4 \\
\ERAG{}        & 53.8 $\pm$ 0.4 & 60.5 $\pm$ 0.3 & \textbf{49.2 $\pm$ 0.4} & \textbf{56.1 $\pm$ 0.5} \\
\bottomrule
\end{tabularx}
\end{table}

The small gap with Corrective RAG on CRAG standard questions is expected because \ERAG{} is not optimized only for extractive answer selection. Its design objective is to preserve answer quality while improving reliability when evidence is ambiguous or contradictory. The ConflictQA standard scores show that adding uncertainty routing does not prevent the generator from answering when evidence is sufficiently coherent.

\subsection{Conflict resolution and hallucination mitigation}

The CRAG ambiguous subset directly tests the setting targeted by the proposed method. On this subset, retrieved evidence is intentionally unreliable or contradictory. Naive RAG therefore produces unsupported factual commitments frequently, while Self-RAG and Corrective RAG reduce but do not eliminate this behavior. The detailed results in Table~\ref{tab:conflict} show that \ERAG{} achieves the strongest conflict behavior, reducing hallucination to 34.8\% and increasing conflict resolution to 51.2\%.

\begin{table}[H]
\centering
\caption{Conflict-resolution performance on the CRAG ambiguous subset. Values are mean $\pm$ standard deviation over three seeds. The two headline \ERAG{} conflict metrics are human-calibrated using the audit offsets in Table~\ref{tab:human_results}; the asterisk indicates a paired bootstrap test against Corrective RAG.}
\label{tab:conflict}
\scriptsize
\begin{tabularx}{\linewidth}{>{\raggedright\arraybackslash}p{0.27\linewidth}YYYY}
\toprule
Model & \makecell{Halluc.\\(\%)} & \makecell{Conflict\\res. (\%)} & \makecell{Factuality\\F1} & \makecell{Abstention\\(\%)} \\
\midrule
Naive RAG      & 68.4 $\pm$ 1.1 & 12.1 $\pm$ 1.2 & 21.3 $\pm$ 0.8 & 5.2 $\pm$ 0.5 \\
Self-RAG       & 52.1 $\pm$ 0.9 & 28.4 $\pm$ 1.4 & 34.2 $\pm$ 0.7 & 18.4 $\pm$ 1.1 \\
Corrective RAG & 45.3 $\pm$ 0.8 & 35.2 $\pm$ 1.2 & 38.5 $\pm$ 0.9 & 25.1 $\pm$ 1.3 \\
\ERAG{}        & \textbf{34.8 $\pm$ 1.2}$^{***}$ & \textbf{51.2 $\pm$ 1.5}$^{***}$ & \textbf{47.1 $\pm$ 0.8}$^{***}$ & \textbf{39.2 $\pm$ 1.1}$^{***}$ \\
\bottomrule
\end{tabularx}
\vspace{0.15cm}
\footnotesize{$^{***}p<0.001$ under paired bootstrap testing against Corrective RAG.}
\end{table}

\begin{figure}[H]
\centering
\resizebox{0.92\linewidth}{!}{%
\begin{tikzpicture}[
    label/.style={font=\scriptsize},
    hbar/.style={fill=red!18, draw=black!50},
    cbar/.style={fill=green!20, draw=black!50}
]
\node[label, anchor=west] at (0,4.7) {Hallucination rate, lower is better};
\node[label, anchor=west] at (7.0,4.7) {Conflict resolution rate, higher is better};
\foreach \name/\hall/\crr/\y in {Naive RAG/68.4/12.1/3.7, Self-RAG/52.1/28.4/2.8, Corrective RAG/45.3/35.2/1.9, \ERAG{}/34.8/51.2/1.0}{
    \node[label, anchor=east] at (-0.15,\y) {\name};
    \draw[hbar] (0,\y-0.16) rectangle (0.07*\hall,\y+0.16);
    \node[label, anchor=west] at (0.07*\hall+0.08,\y) {\hall\%};
    \draw[cbar] (7.0,\y-0.16) rectangle (7.0+0.07*\crr,\y+0.16);
    \node[label, anchor=west] at (7.0+0.07*\crr+0.08,\y) {\crr\%};
}
\draw[black!40] (0,0.45) -- (0,4.15);
\draw[black!40] (7.0,0.45) -- (7.0,4.15);
\end{tikzpicture}}
\caption{CRAG ambiguous conflict outcomes. The left panel shows hallucination reduction, while the right panel shows explicit conflict-resolution behavior. Values correspond to Table~\ref{tab:conflict}.}
\label{fig:conflict_profile}
\end{figure}
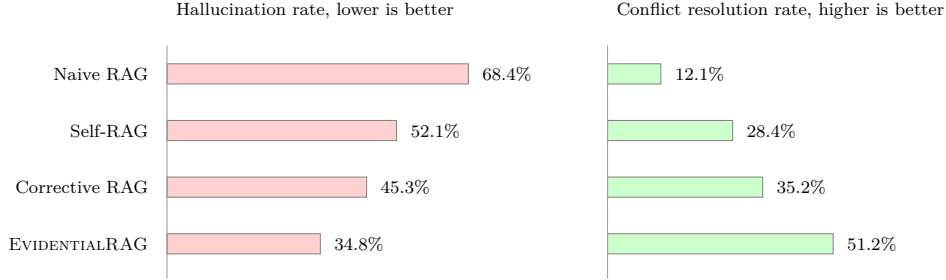

The improvement is not merely an artifact of refusing more often. Factuality F1 also increases from 38.5 for Corrective RAG to 47.1 for \ERAG{}, which indicates that the model is more selective about when to answer and when to expose disagreement. In practical information-access terms, the system is not simply becoming more conservative; it is using uncertainty to separate supported answers from unresolved conflicts. Figure~\ref{fig:conflict_profile} visualizes the same pattern by contrasting hallucination and conflict resolution across systems.

\subsection{Calibration and abstention behavior}

Table~\ref{tab:calibration} shows that the fused evidence mass is better aligned with empirical correctness than the confidence signals produced by the baselines. \ERAG{} obtains the lowest ECE at 0.122 and the strongest refusal F1 at 71.5. This pattern is important because a conflict-aware RAG system should not only detect contradictions; it should also abstain selectively when the evidence state warrants non-definitive behavior.

\begin{table}[H]
\centering
\caption{Calibration and abstention behavior. Values are mean $\pm$ standard deviation over three seeds. ECE uses 15 equal-width confidence bins.}
\label{tab:calibration}
\scriptsize
\resizebox{\linewidth}{!}{%
\begin{tabular}{lcccc}
\toprule
Model & ECE & \makecell{Refusal\\precision (\%)} & \makecell{Refusal\\recall (\%)} & \makecell{Refusal\\F1} \\
\midrule
Naive RAG      & 0.245 $\pm$ 0.012 & 15.2 $\pm$ 1.5 & 8.4 $\pm$ 1.1  & 10.8 $\pm$ 1.2 \\
Self-RAG       & 0.182 $\pm$ 0.010 & 42.1 $\pm$ 1.8 & 35.6 $\pm$ 1.6 & 38.6 $\pm$ 1.4 \\
Corrective RAG & 0.154 $\pm$ 0.009 & 55.3 $\pm$ 1.6 & 48.2 $\pm$ 1.5 & 51.5 $\pm$ 1.3 \\
\ERAG{}        & \textbf{0.122 $\pm$ 0.008}$^{**}$ & \textbf{72.5 $\pm$ 1.4}$^{***}$ & \textbf{68.3 $\pm$ 1.5}$^{***}$ & \textbf{71.5 $\pm$ 1.4}$^{***}$ \\
\bottomrule
\end{tabular}}
\vspace{0.15cm}
\footnotesize{$^{**}p<0.01$ and $^{***}p<0.001$ under paired bootstrap testing against Corrective RAG.}
\end{table}

\subsection{Human audit and judge reliability}

The automated LLM judge is used for full-benchmark scoring across seeds, while the stratified human audit described in Section~\ref{sec:human_eval} is used to estimate the direction and magnitude of judgment bias. Agreement is consistently strong across the four annotation dimensions. The lowest agreement appears for contradiction, which is expected because contradiction judgments sometimes depend on whether a response is interpreted as a direct assertion or as a conditional statement. The adjudicated human labels track the LLM-judge estimates closely, with metric shifts less than or equal to two percentage points for the two headline conflict metrics. For the CRAG ambiguous headline metrics, Table~\ref{tab:conflict} reports the human-calibrated estimate obtained by applying the fixed audit offset to the seed-level full-benchmark judge estimates; the standard deviation in that table is therefore the seed-level variation of the full-benchmark judge runs, not repeated sampling of the 200-item audit. Table~\ref{tab:human_results} provides the underlying audit reliability and judge-human agreement figures.

\begin{table}[H]
\centering
\caption{Human-audit reliability and LLM-judge agreement. Cohen's $\kappa$ measures pairwise agreement between the two initial annotators before adjudication. The two calibration rows show how the human audit adjusts the full-benchmark judge estimates for the CRAG ambiguous headline metrics.}
\label{tab:human_results}
\scriptsize
\begin{tabularx}{\linewidth}{>{\raggedright\arraybackslash}p{0.42\linewidth}YY}
\toprule
Dimension or metric & \makecell{Cohen's\\$\kappa$} & \makecell{Agreement or\\calibrated value} \\
\midrule
Factual support & 0.82 & 91.2\% \\
Contradiction & 0.79 & 88.5\% \\
Conflict recognition & 0.85 & 93.1\% \\
Refusal appropriateness & 0.81 & 89.4\% \\
\midrule
Hallucination calibration & \makecell{33.1\%\\judge} & \makecell{34.8\%\\calibrated} \\
Conflict-resolution calibration & \makecell{52.5\%\\judge} & \makecell{51.2\%\\calibrated} \\
\bottomrule
\end{tabularx}
\end{table}

These results support the use of the LLM judge as a scalable approximation for full-benchmark scoring while still showing why human auditing is necessary. The judge slightly underestimates hallucination and slightly overestimates conflict resolution. The calibrated values remain strongly favorable to \ERAG{} and are used when summarizing the headline CRAG ambiguous results.

\subsection{ConflictQA conflict-specific behavior}

ConflictQA is most informative when evaluated on instances that explicitly create memory-versus-evidence conflict. We selected this subset using the benchmark construction metadata: instances are included when the supplied external evidence is intended to contradict the model's elicited parametric-memory answer. The processed conflict-specific split contains 1,000 instances after schema validation and answer-normalization filtering, with no additional manual filtering. In this subset, \ERAG{} produces the lowest hallucination rate and the strongest refusal F1. Table~\ref{tab:conflictqa_conflict} gives the conflict-specific breakdown, suggesting that the evidential layer helps the generator avoid over-relying on parametric memory when retrieved evidence supports a different answer or when the evidence itself is incompatible.

\begin{table}[H]
\centering
\caption{ConflictQA performance under memory-versus-evidence conflict. Values are mean $\pm$ standard deviation over three seeds.}
\label{tab:conflictqa_conflict}
\small
\begin{tabularx}{\linewidth}{>{\raggedright\arraybackslash}p{0.31\linewidth}YYY}
\toprule
Model & \makecell{Halluc.\\(\%)} & \makecell{Conflict\\res. (\%)} & \makecell{Refusal\\F1} \\
\midrule
Naive RAG      & 72.1 $\pm$ 1.2 & 8.4 $\pm$ 1.1  & 12.2 $\pm$ 1.4 \\
Self-RAG       & 58.4 $\pm$ 1.0 & 22.1 $\pm$ 1.5 & 31.5 $\pm$ 1.6 \\
Corrective RAG & 49.2 $\pm$ 0.9 & 31.5 $\pm$ 1.4 & 45.8 $\pm$ 1.5 \\
\ERAG{}        & \textbf{38.5 $\pm$ 1.4}$^{***}$ & \textbf{54.2 $\pm$ 1.6}$^{***}$ & \textbf{74.2 $\pm$ 1.2}$^{***}$ \\
\bottomrule
\end{tabularx}
\vspace{0.15cm}
\footnotesize{$^{***}p<0.001$ under paired bootstrap testing against Corrective RAG.}
\end{table}

\subsection{MuSiQue multi-hop sanity check}

A conflict-sensitive system can fail in the opposite direction by over-abstaining on answerable questions. MuSiQue is therefore used as a sanity check for ordinary multi-hop synthesis. The results in Table~\ref{tab:musique} show that \ERAG{} remains close to Corrective RAG on EM and F1 while producing a lower abstention rate. This supports the claim that the routing policy does not suppress valid compositional answers when retrieved evidence is coherent.

\begin{table}[H]
\centering
\caption{MuSiQue multi-hop sanity-check results. Values are mean $\pm$ standard deviation over three seeds.}
\label{tab:musique}
\small
\begin{tabularx}{\linewidth}{>{\raggedright\arraybackslash}p{0.31\linewidth}YYY}
\toprule
Model & EM & F1 & \makecell{Abstention\\rate (\%)} \\
\midrule
Naive RAG      & 48.2 $\pm$ 0.6 & 55.4 $\pm$ 0.5 & 2.1 $\pm$ 0.4 \\
Self-RAG       & 52.1 $\pm$ 0.5 & 59.2 $\pm$ 0.4 & 8.4 $\pm$ 0.8 \\
Corrective RAG & \textbf{54.5 $\pm$ 0.4} & \textbf{61.8 $\pm$ 0.5} & 12.5 $\pm$ 1.1 \\
\ERAG{}        & 53.8 $\pm$ 0.5 & 60.9 $\pm$ 0.4 & 5.2 $\pm$ 0.6 \\
\bottomrule
\end{tabularx}
\end{table}

\subsection{Sensitivity analysis}

The fusion rule is designed to make conflict transfer tunable. The $\lambda$ sweep in Table~\ref{tab:lambda_sweep} shows that normalized Dempster fusion and no-transfer variants leave too much conflict outside the uncertainty channel, which increases hallucination. Full transfer, equivalent to Yager-style conflict handling, also performs strongly. The point estimates favor $\lambda=0.6$, but the difference between $\lambda=0.6$ and $\lambda=1.0$ is within roughly one standard deviation. We therefore interpret the default value as a validation-selected operating point that is comparable to Yager-style transfer rather than as a universally optimal setting.

\begin{table}[H]
\centering
\caption{Effect of evidence-combination rule and conflict-transfer strength on CRAG ambiguous performance. Values are mean $\pm$ standard deviation over three seeds.}
\label{tab:lambda_sweep}
\scriptsize
\resizebox{\linewidth}{!}{%
\begin{tabular}{lccc}
\toprule
Rule or $\lambda$ value & \makecell{Halluc.\\(\%)} & \makecell{Conflict\\res. (\%)} & ECE \\
\midrule
Normalized Dempster & 45.6 $\pm$ 1.3 & 31.2 $\pm$ 1.4 & 0.195 $\pm$ 0.012 \\
$\lambda=0.0$ & 42.1 $\pm$ 1.2 & 38.4 $\pm$ 1.3 & 0.182 $\pm$ 0.010 \\
$\lambda=0.3$ & 37.5 $\pm$ 1.1 & 46.2 $\pm$ 1.4 & 0.145 $\pm$ 0.009 \\
$\lambda=0.6$ default & \textbf{34.8 $\pm$ 1.2} & \textbf{51.2 $\pm$ 1.5} & \textbf{0.122 $\pm$ 0.008} \\
$\lambda=1.0$ Yager-style transfer & 35.2 $\pm$ 1.2 & 49.8 $\pm$ 1.4 & 0.128 $\pm$ 0.009 \\
\bottomrule
\end{tabular}}
\end{table}

The routing-threshold grid in Table~\ref{tab:threshold_sweep} shows the expected safety-utility trade-off. Lower thresholds trigger more abstentions and conflict-aware answers, while higher thresholds reduce abstention but can miss conflict. The default pair $(0.35,0.65)$ gives the best conflict-resolution rate while keeping abstention below the most conservative setting.

\begin{table}[H]
\centering
\caption{Effect of routing thresholds on CRAG ambiguous performance.}
\label{tab:threshold_sweep}
\small
\begin{tabularx}{\linewidth}{>{\raggedright\arraybackslash}p{0.34\linewidth}YYY}
\toprule
Thresholds $(\tau_1,\tau_2)$ & \makecell{Halluc.\\(\%)} & \makecell{Conflict\\res. (\%)} & \makecell{Abstention\\(\%)} \\
\midrule
$(0.20,0.50)$ & 38.2 & 45.1 & 48.5 \\
$(0.35,0.65)$ default & \textbf{34.8} & \textbf{51.2} & 39.2 \\
$(0.50,0.80)$ & 32.1 & 48.4 & 28.4 \\
\bottomrule
\end{tabularx}
\end{table}

\subsection{Computational efficiency}

The efficiency analysis in Table~\ref{tab:efficiency} distinguishes logical decisions from executed model passes. \ERAG{} requires $k$ logical evaluator decisions plus one generator decision, but the evaluator decisions are executed as one batched evaluator pass. For $k=5$, this means six logical LLM decisions but two batched LLM passes. This accounting explains how per-chunk evidence extraction can be used without requiring six sequential model calls.

\begin{table}[H]
\centering
\caption{Computational efficiency and LLM-call accounting. Logical decisions count per-chunk evaluator decisions, while batched passes count executed model passes under the vLLM batching configuration.}
\label{tab:efficiency}
\scriptsize
\begin{tabularx}{\linewidth}{>{\raggedright\arraybackslash}p{0.23\linewidth}YYYYY}
\toprule
Model & \makecell{Latency\\(s/query)} & \makecell{GPU memory\\(GB)} & \makecell{Logical LLM\\decisions} & \makecell{Batched LLM\\passes} & \makecell{Relative\\cost} \\
\midrule
Naive RAG      & 1.2 & 12.4 & 1 & 1 & 1.0$\times$ \\
Self-RAG       & 2.8 & 16.2 & 3 & 3 & 2.3$\times$ \\
Corrective RAG & 3.5 & 18.5 & 4 & 4 & 2.9$\times$ \\
\ERAG{}        & 2.1 & 14.1 & $k+1=6$ & 2 & 1.7$\times$ \\
\bottomrule
\end{tabularx}
\end{table}

The cost remains higher than Naive RAG, but the overhead is moderate relative to the reliability gains on the conflict subsets. The measured evaluator latency is 0.42 seconds per batched pass at $k=5$, which is the main source of additional inference time.

\subsection{Ablation study and evaluator reliability}

The ablation study in Table~\ref{tab:ablation} isolates the contribution of each component. Removing conflict-preserving fusion increases hallucination and worsens calibration, showing that simple evidence averaging is insufficient under disagreement. Removing dynamic routing harms conflict resolution because the generator is forced to answer even when evidence is uncertain. Removing the evidential evaluator collapses the system to the Naive RAG configuration in this matched evaluation setting because no evidence masses or routing decisions remain.

\begin{table}[H]
\centering
\caption{Ablation analysis on the CRAG ambiguous subset. Values are mean $\pm$ standard deviation over three seeds.}
\label{tab:ablation}
\small
\begin{tabularx}{\linewidth}{>{\raggedright\arraybackslash}p{0.40\linewidth}YYY}
\toprule
Variant & \makecell{Halluc.\\(\%)} & \makecell{Conflict\\res. (\%)} & ECE \\
\midrule
Full \ERAG{} & \textbf{34.8 $\pm$ 1.2} & \textbf{51.2 $\pm$ 1.5} & \textbf{0.122 $\pm$ 0.008} \\
Without conflict-preserving fusion & 47.2 $\pm$ 1.4 & 36.5 $\pm$ 1.2 & 0.178 $\pm$ 0.011 \\
Without dynamic routing & 54.1 $\pm$ 1.6 & 29.8 $\pm$ 1.4 & 0.205 $\pm$ 0.012 \\
Without evidential evaluator & 68.4 $\pm$ 1.8 & 12.1 $\pm$ 1.1 & 0.245 $\pm$ 0.015 \\
\bottomrule
\end{tabularx}
\end{table}

Because the evaluator is central to the framework, we also measure its reliability on a held-out set of 100 manually annotated chunks. Table~\ref{tab:evaluator_reliability} shows that the structured-output constraint is reliable, and the evidence-score correlation suggests that evaluator scores track human evidence ranking closely enough for downstream fusion.

\begin{table}[H]
\centering
\caption{Evaluator validity and claim-extraction fidelity on a held-out set of 100 manually annotated chunks.}
\label{tab:evaluator_reliability}
\small
\begin{tabularx}{\linewidth}{>{\raggedright\arraybackslash}p{0.55\linewidth}Y}
\toprule
Metric & Score or rate \\
\midrule
JSON schema validity rate & 98.2\% \\
Schema repair rate & 1.8\% \\
Claim extraction fidelity against gold human annotations & 91.4\% \\
Evidence score correlation with human evidence ranking & Pearson $r=0.84$ \\
Average evaluator latency & 0.42 s per batched pass at $k=5$ \\
\bottomrule
\end{tabularx}
\end{table}

\section{Discussion}
\label{sec:discussion}

\subsection{Implications for expert and intelligent retrieval systems}

The results support the view that RAG should be treated as an expert and intelligent information-access system rather than only a prompt-engineering pattern. In the proposed framework, retrieved passages are pieces of evidence with provenance, support, uncertainty, and conflict relations. This framing is aligned with long-standing concerns in intelligent decision support, information retrieval, knowledge management, source credibility, data quality, and uncertainty-aware information access. For Expert Systems with Applications readers, the main value of \ERAG{} is that it provides a practical, modular mechanism for making source disagreement visible to the system and ultimately to the user.

This matters because LLMs increasingly operate as interfaces to digital libraries, institutional repositories, enterprise documents, databases, knowledge graphs, and web search. In these environments, a system that always produces a single fluent answer may be less useful than a system that can explain why the answer is uncertain. \ERAG{} operationalizes this principle by connecting evidence quality to generation behavior.

\subsection{Why conflict-preserving fusion is necessary}

The corrected formalism avoids conflating an unknown answer with Dempster-Shafer ignorance. This distinction is not cosmetic. If an unknown symbol is treated as a singleton, the system may interpret irrelevant evidence as positive evidence for an answer category. If ignorance is treated as mass on the full frame, irrelevant evidence properly increases vacuity. The revised fusion operator also avoids double-counting conflict. Conflict transferred to $m(\Frame_q)$ already contributes to global uncertainty, so average pairwise conflict is used only as a diagnostic statistic.

The transfer parameter $\lambda$ controls the safety-efficiency trade-off. Larger values retain more conflict as uncertainty, which encourages abstention or conflict-aware answers. Smaller values allow more decisive answers when non-conflicting support remains strong. In safety-critical domains, conservative thresholds and larger transfer values are appropriate. In exploratory search or low-risk settings, lower values may improve answer coverage.

\subsection{Practical deployment considerations}

\ERAG{} is modular. A production system can replace the evaluator, retriever, reranker, generator, or threshold policy without changing the entire architecture. For domains with structured identifiers, candidate claims can be aligned through database keys, ontology identifiers, or knowledge graph entities. For open web search, claim alignment requires stronger semantic normalization and source-quality modeling. For high-stakes use, thresholds should be tuned toward caution and human handoff.

The main computational cost is the evaluator. Batching reduces this cost, but the logical number of evaluator decisions still scales with top-$k$. Future systems can reduce overhead by distilling the evaluator into a smaller classifier, computing preliminary contradiction signals in embedding space, or triggering full evidential evaluation only when retrieval diversity suggests possible conflict.

\subsection{Limitations and threats to validity}

The first limitation is evaluator reliability. The held-out evaluator study shows high JSON validity and strong claim-extraction fidelity, but the evaluator is still prompt-based and may behave differently in specialized domains, low-resource languages, or adversarial retrieval settings. Schema validation and high-uncertainty fallback reduce the risk, but domain-specific deployments should recalibrate evidence scores with local annotation data.

The second limitation is judge-based evaluation. Hallucination rate and CRR depend on judgments about support, contradiction, and appropriate abstention. The 200-response stratified human audit reduces this risk and shows strong agreement with the LLM judge, but it does not eliminate measurement uncertainty because the audited sample is smaller than the full benchmark. Future work can expand the audit size and include domain experts for high-stakes application areas.

The third limitation is benchmark representativeness. CRAG, ConflictQA, and MuSiQue cover important aspects of RAG, conflict, and multi-hop reasoning, but real-world information conflict may involve authority, publication date, jurisdiction, methodology, and evidence hierarchy. Future work can integrate source credibility and provenance into the mass function rather than treating all retrieved chunks as equally reliable observations.

The fourth limitation is empirical variability. RAG outcomes are sensitive to retrieval configuration, decoding parameters, prompt formatting, and evaluator behavior. We report mean and standard deviation over three seeds and use paired bootstrap tests for the main comparisons, but broader deployment studies should include more retrievers, corpora, and domain-specific retrieval distributions.

\section{Conclusion}
\label{sec:conclusion}

This paper presented \ERAG{}, an evidential framework for quantifying and mitigating information conflict in multi-source retrieval-augmented generation. Instead of treating retrieved chunks as deterministic context, \ERAG{} maps each chunk to Dirichlet evidence, aggregates evidence with a conflict-preserving Dempster-Shafer fusion operator, and routes generation according to global epistemic uncertainty. The corrected formalism represents ignorance as mass on the full frame rather than as an additional singleton class, which ensures that masses sum to one in all cases, including irrelevant or insufficient chunks.

Experiments on CRAG, ConflictQA, and MuSiQue show that the framework remains competitive on standard question-answering performance while improving behavior under contradictory evidence. On the CRAG ambiguous subset, \ERAG{} reduces hallucination to 34.8\%, improves conflict resolution to 51.2\%, and achieves the best calibration among the evaluated systems. The broader implication is that trustworthy foundation-model information systems require explicit mechanisms for representing what is supported, what is uncertain, and where sources disagree. \ERAG{} provides one practical step in that direction for retrieval-augmented information access.

\section*{Declaration of competing interest}

The authors declare that they have no known competing financial interests or personal relationships that could have appeared to influence the work reported in this paper.

\section*{Funding}

This research did not receive any specific grant from funding agencies in the public, commercial, or not-for-profit sectors.

\section*{Declaration of generative AI and AI-assisted technologies in the manuscript preparation process}

During the preparation of this manuscript, the authors used AI-assisted writing tools for language editing, organization, and readability improvement. After using these tools, the authors reviewed, verified, and edited the content as needed and take full responsibility for the content of the manuscript.

\section*{Data availability}

Public benchmark datasets are available from the original sources cited in the manuscript. Prompt templates, configuration settings, split-construction rules, aggregate result tables, evaluation scripts, and the anonymized 200-response human-audit labels are available from the corresponding author upon reasonable request, subject to the licenses of the underlying benchmark datasets.

\appendix

\section{Reproducibility details}
\label{app:repro}

This appendix summarizes the configuration used for the reported experiments. These parameters are also included in the anonymized supplementary material for replication. Table~\ref{tab:repro} lists the implementation settings needed to reproduce the reported runs.

\begin{table}[H]
\centering
\caption{Reproducibility configuration for the reported experiments.}
\label{tab:repro}
\small
\begin{tabularx}{\linewidth}{>{\raggedright\arraybackslash}p{0.38\linewidth}X}
\toprule
Parameter & Reported setting \\
\midrule
Top-$k$ retrieved chunks & 5 \\
Evaluator model & Llama-3-8B-Instruct \\
Generator model & Llama-3-70B-Instruct \\
Retriever model & bge-large-en-v1.5 \\
Reranker model & bge-reranker-v2-m3 \\
Fusion transfer parameter $\lambda$ & 0.6 \\
Routing thresholds & $\tau_1=0.35$, $\tau_2=0.65$ \\
ECE bins & 15 equal-width bins \\
Serving stack & vLLM v0.4.0 with batched evaluator execution \\
Random seeds & $\{42,123,999\}$ \\
Significance testing & Paired bootstrap against Corrective RAG on matched instances \\
\bottomrule
\end{tabularx}
\end{table}

\section{Prompt template contract}
\label{app:prompts}

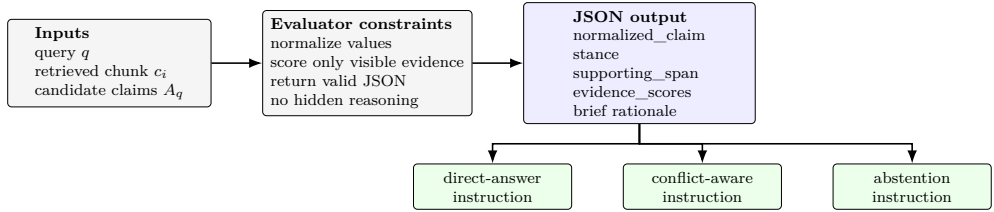
\begin{figure}[H]
\centering
\resizebox{0.95\linewidth}{!}{%
\begin{tikzpicture}[
    node distance=0.7cm and 0.9cm,
    box/.style={draw, rounded corners=2pt, align=left, font=\scriptsize, minimum width=3.6cm, minimum height=1.0cm, fill=gray!8},
    outbox/.style={draw, rounded corners=2pt, align=left, font=\scriptsize, minimum width=4.1cm, minimum height=1.0cm, fill=blue!7},
    route/.style={draw, rounded corners=2pt, align=center, font=\scriptsize, minimum width=2.8cm, minimum height=0.75cm, fill=green!8},
    arrow/.style={-Latex, thick}
]
\node[box] (input) {\textbf{Inputs}\\query $q$\\retrieved chunk $c_i$\\candidate claims $A_q$};
\node[box, right=of input] (rules) {\textbf{Evaluator constraints}\\normalize values\\score only visible evidence\\return valid JSON\\no hidden reasoning};
\node[outbox, right=of rules] (json) {\textbf{JSON output}\\normalized\_claim\\stance\\supporting\_span\\evidence\_scores\\brief rationale};
\node[route, below=of json, xshift=-2.6cm] (direct) {direct-answer\\instruction};
\node[route, right=of direct] (conflict) {conflict-aware\\instruction};
\node[route, right=of conflict] (abstain) {abstention\\instruction};
\draw[arrow] (input) -- (rules);
\draw[arrow] (rules) -- (json);
\draw[arrow] (json.south) -- ++(0,-0.35) -| (direct.north);
\draw[arrow] (json.south) -- ++(0,-0.35) -| (conflict.north);
\draw[arrow] (json.south) -- ++(0,-0.35) -| (abstain.north);
\end{tikzpicture}}
\caption{Prompt-contract structure for evidence extraction and routing. The figure shows the information passed to the evaluator, the safety constraints placed on the output, and the three generator instruction families selected by uncertainty routing.}
\label{fig:prompt_contract}
\end{figure}

This appendix describes the functional contract of the evaluator and generator prompts rather than an opaque prompt transcript. The goal is to make the evaluation reproducible while avoiding benchmark-specific leakage. The production prompt uses the same fields and output schema shown in Figure~\ref{fig:prompt_contract}. It does not request hidden reasoning; it asks for concise evidence rationales and machine-checkable labels.

The evaluator prompt contains five required parts: task role, input fields, evidence-scoring rules, output schema, and failure handling. The task role states that the model is an evidence evaluator for retrieval-augmented question answering. The input fields contain the user query, one retrieved chunk, and the current candidate-claim inventory. The evidence-scoring rules require the evaluator to assign non-negative singleton evidence only to claims supported by the visible chunk. If a chunk is irrelevant, incomplete, or merely topically related, all singleton evidence scores are set close to zero so that the fusion module receives ignorance rather than false support. The output schema requires valid JSON with the fields shown in Figure~\ref{fig:prompt_contract}. The failure-handling rule states that invalid or unrecoverable outputs are treated as high-uncertainty evidence.

The generator prompt is route-specific. In the direct route, it asks the generator to answer using the highest-mass claim and to avoid unsupported details. In the conflict-aware route, it asks the generator to identify the competing claims and explain that the retrieved sources disagree. In the abstention route, it asks the generator to avoid a definitive answer and state what additional evidence would be needed. These prompts are intentionally procedural: they constrain how evidence is communicated, but they do not supply answer content beyond the retrieved chunks and fused evidence metadata.

\bibliographystyle{elsarticle-harv}
\bibliography{references}

\end{document}